\documentclass[letterpaper,10pt]{article} 

\usepackage{opticameet3} 

\usepackage{amsmath,amssymb}
\usepackage[colorlinks=true,bookmarks=false,citecolor=blue,urlcolor=blue]{hyperref} 
\usepackage{upgreek}
\usepackage{multirow}
\usepackage{pgfplots}
\pgfplotsset{compat=1.9}
\usepackage{graphicx}
\usepackage{subcaption}
\usepackage{wrapfig, blindtext}
\usepackage{mwe}
\usepackage{adjustbox}
\usepackage{xcolor}
\usepackage{collcell}
\usepackage{mathtools}
\usepackage{colortbl}%
  
\usepackage[colorlinks=true,bookmarks=false,citecolor=blue,urlcolor=blue]{hyperref} 
\usepackage{float}
\floatstyle{plaintop}
\restylefloat{table}

\begin{document}


\title{Hardware Realization of Nonlinear Activation Functions for NN-based Optical Equalizers}

\author{Sasipim Srivallapanondh\textsuperscript{1,*}, Pedro J. Freire\textsuperscript{1}, Antonio Napoli\textsuperscript{2}, \\Sergei K. Turitsyn\textsuperscript{1}, Jaroslaw E. Prilepsky \textsuperscript{1}}
\address{\textsuperscript{1}Aston University, B4 7ET Birmingham, UK;  \;\;\textsuperscript{2}Infinera, Munich, Germany}
\email{*s.srivallapanondh@aston.ac.uk}

\copyrightyear{2022}

\begin{abstract}
    To reduce the complexity of the hardware implementation of neural network-based optical channel equalizers, we demonstrate that the performance of the biLSTM equalizer with approximated activation functions is close to that of the original model.
\end{abstract}

\vspace{-0.1mm}
\section{Introduction}
\vspace{-1mm}
The information rate in coherent optical transmission systems is limited by nonlinear optical fiber impairments. Thus, different digital signal processing (DSP) techniques have been proposed to alleviate nonlinear effects \cite{cartledge2017digital}. More recently, the research focus has shifted to the deployment of neural networks (NNs) for optical channel post-equalization, because NNs have shown the capability to estimate the inverse optical channel transfer function and successfully mitigate the nonlinearity. However, major challenges still prevent NN-based equalizers from being implemented in real hardware, in particular, the high computational complexity \cite{Pedro2021,pedro_fpga,tommiska2003efficient}. The complexity of nonlinear activation functions is one of the key components when designing the NNs in hardware. In the hardware realization, the weights and inputs of the NN can be converted directly from the float- to fixed-point representation, while the nonlinear activation function's realization is more challenging. Particularly, the bidirectional long-short term memory (biLSTM)-based equalizer, which shows the best performance in mitigating nonlinear impairments \cite{Pedro2021,pedro_fpga}, incorporates \textit{sigmoid} and \textit{tanh} as its activation functions. These functions are computationally expensive as they contain exponential expressions, which require large chip areas and are not suitable for resource-constrained hardware \cite{tommiska2003efficient,ccetin2015application}. Consequently, the approximation methods for these nonlinear functions are necessary to reduce the computational complexity and allow hardware realization of the NN\cite{li2022fpga,ccetin2015application,basterretxea2001approximation}. The popular techniques for the nonlinear function approximations are: Taylor series expansion, piecewise linear (PWL), and lookup tables (LUT). The PWL draws our attention because this method consumes fewer resources compared to the Taylor expansion that fits the functions with high-order polynomials, while the LUT requires too much storage space to save high-precision values if we aim to achieve high accuracy\cite{li2022fpga}. 

In this work, we focus on the sigmoid and tanh approximations by using the PWL technique. We also investigate the scenario in which the training of the NN with approximated activation functions is undertaken to reduce the approximation errors. Finally, we evaluate the amount of resources required, in terms of LUT, flipflops (FF), and DSP slices, within the field programmable gate arrays (FPGA) implementation by using the PWL approximation.

\vspace{-2mm}
\section{Nonlinear Activation Function Approximation: Performance versus Complexity Trade-off Using PWL}
\begin{figure*}[t!]
  \centering
\begin{subfigure}{.33\textwidth}
  \centering
    \includegraphics[scale=0.26]{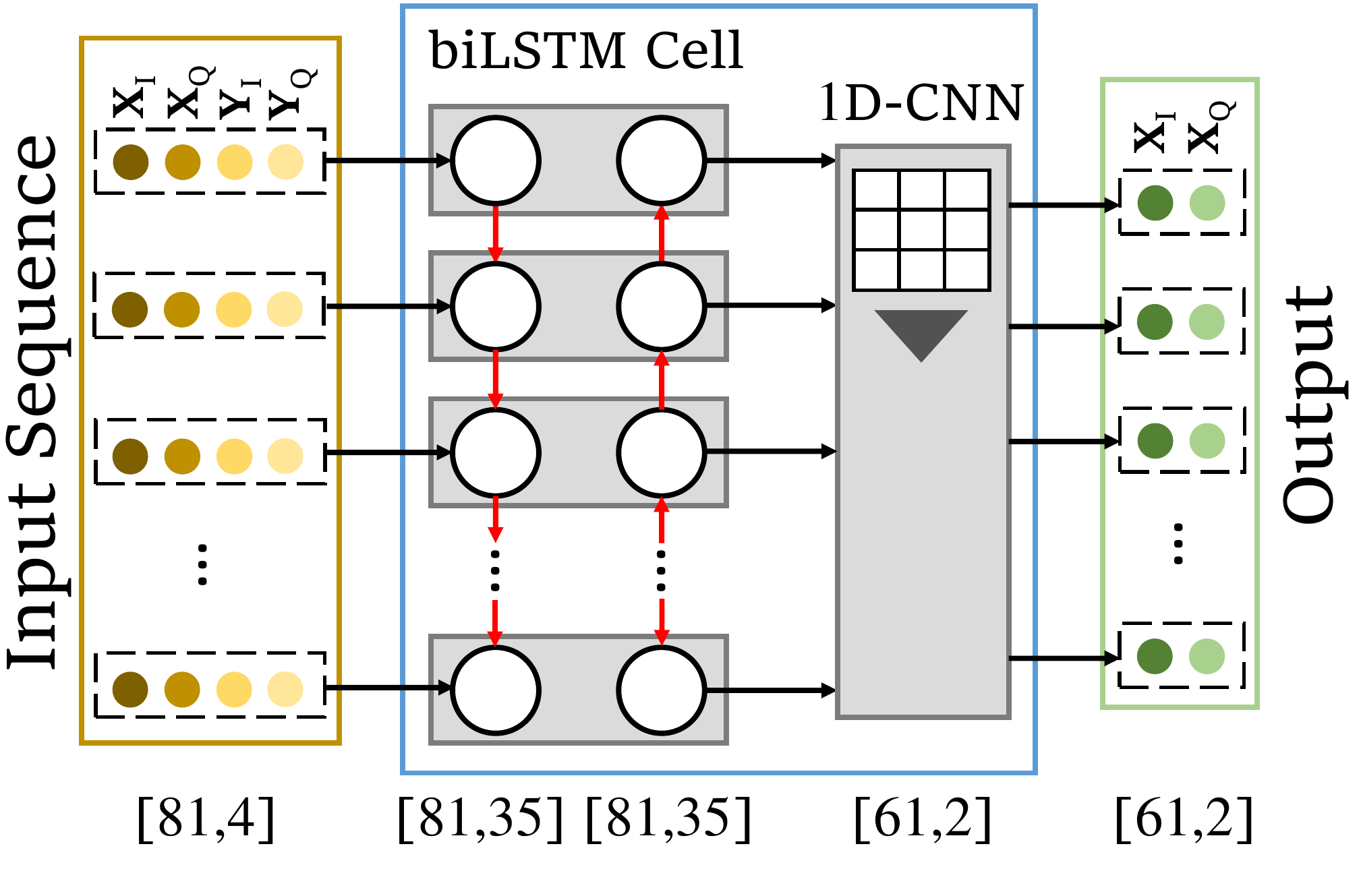}
    \put(-157, 96){\textcolor{black}{(a)}}
\end{subfigure}
\hfill
\begin{subfigure}{.33\textwidth}
  \centering  
  \includegraphics[scale=0.26]{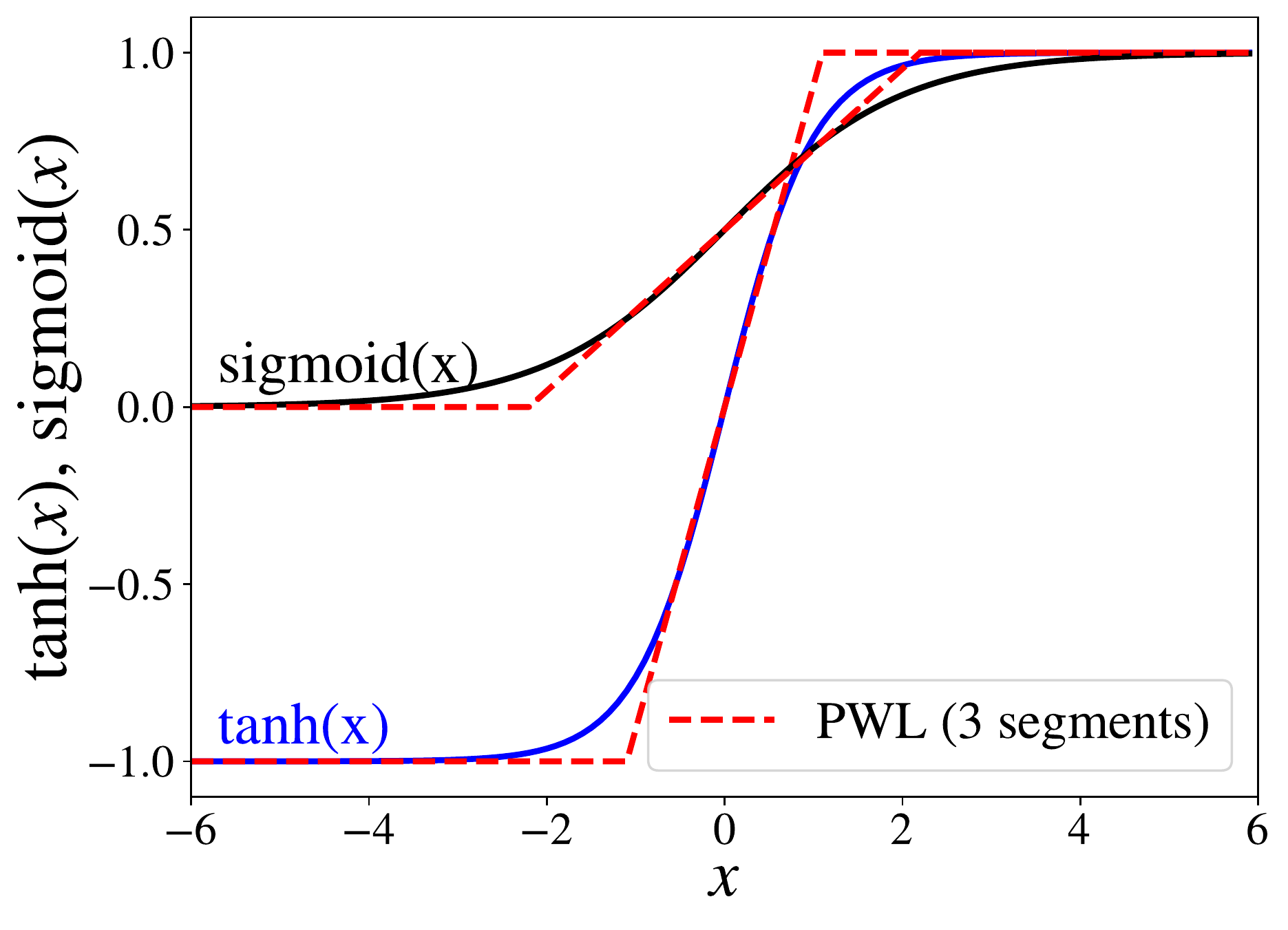}
  \put(-140,96){\textcolor{black}{(b)}}
\end{subfigure}
\hfill
\begin{subfigure}{.30\textwidth}
  \centering
  \begin{tikzpicture}[scale=0.537]
    \begin{axis} [
        xlabel={No. of segments},
        ylabel={Q-Factor [dB]},
        grid=both,  
        ylabel near ticks,
        xmin=3, xmax=9,
    	xtick={3,5,7,9},
    	ymin=0, ymax=5.5,
        legend style={legend pos=south east, legend cell align=left,fill=white, fill opacity=0.6, draw opacity=1,text opacity=1},
    	grid style={dashed}]
        ]
        \addplot[color=green, very thick]     coordinates {
    (2, 5.2)(16, 5.2)
    };
    \addlegendentry{Without approximation};
        \addplot[color=red, mark=*, dashed, very thick]     coordinates {
    (3, 0)(5, 3.21)(7, 4.26)(9, 4.48)
    };
    \addlegendentry{PWL without re-training};
    
    \addplot[color=red, mark=*, very thick]     coordinates {
    (3, 5.09)(5, 5.075)(7, 5.1)(9, 5.1)
    };
    \addlegendentry{PWL with re-training};
        \end{axis}
    \node at (-0.8,5.4) {(c)};
\end{tikzpicture}     
\end{subfigure}
\vspace{-3mm}
\caption{(a) biLSTM+CNN equalizer structure; (b) 3-segment PWL approximation of tanh and sigmoid; (c) performance vs. number of segments used in PWL, considering with and without re-training the weights.}
\label{fig:fig1}
\vspace{-8mm}
\end{figure*}
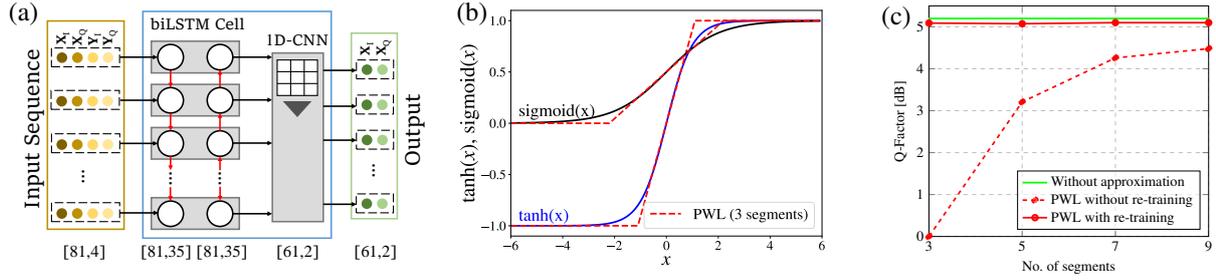

\vspace{-1mm}
The PWL approximation uses a combination of linear segments to fit the nonlinear function \cite{basterretxea2001approximation, li2022fpga}. The higher number of linear segments to represent the nonlinear function yields better accuracy. The PWL approximation is a promising method to allow high-speed processing\footnote{The implementation of PWL can be optimized further to have zero multipliers by simplifying the shift and addition operations\cite{tommiska2003efficient}.} because this technique can provide a comparable level of accuracy to the model with the original activation functions by deploying more segments. The PWL still consumes efficient memory \cite{li2022fpga} since only the coefficients of each linear segment are stored in the memory. When using not enough segments of PWL to approximate the activation functions, the performance of the NN can drastically degrade. 
To reduce the approximation errors, the model with approximated activation functions can be further trained to enhance performance. Our equalizer \cite{pedro_fpga}, as depicted in Fig.~\ref{fig:fig1}a, contains a biLSTM layer with 35 hidden units and a linear 1D-convolutional NN (CNN) layer with a kernel size of 21 without padding and 2 filters to recover real and imaginary parts in its output. The biLSTM+CNN equalizer takes 81 symbols as inputs and retrieves 61 equalized symbols at the output. This equalizer is pre-trained with the actual activation functions which provide the best performance. The NN's activation functions (sigmoid and tanh), then, are replaced by the PWL approximations; see Fig.~\ref{fig:fig1}b as an example of PWL with 3 segments in approximating sigmoid and tanh, also known as ``\textit{hard sigmoid}'' and ``\textit{hard tanh}''. To alleviate the approximation errors, we re-train the weights of the NN. Note that the training can also be carried out from scratch, meaning that the NN is trained when the activation functions are replaced by the approximations from the beginning without any pre-assigned weights. However, training from scratch takes a considerably longer time to achieve the same level of performance as that of the re-training method.

\vspace{-2mm}
\section{Results and Conclusions}
\vspace{-1mm}
We evaluate the Q-factor of our biLSTM+CNN equalizer when applying the 3-, 5-, 7-, and 9- segment PWL approximations to both tanh and sigmoid, shown in Fig.~\ref{fig:fig1}c. Without re-training, the weights after the approximations, the NN equalizer performs progressively better as the number of segments increases; it is a trade-off between complexity and performance. However, when re-training the NN weights after the approximations, an increase in the approximation order barely improves the Q-factor. The PWL with re-training shows a noticeable performance gain and provides comparable performance to that of the NN without activation function approximations, with only a slight 0.1~dB drop in Q-factor. Fig.~\ref{fig:fig1}c highlights that the training can mitigate approximation errors and shows that even low-complexity versions of the PWL can yield a similar performance level to the NN with original activation functions. 
In FPGA, the resources in terms of LUT, FF, and DSP slices are used to build the logic behind the functionality of the approximation. In this work, the hardware realization of the tanh function is undertaken on the EK-VCK190-G-ED Xilinx FPGA chip. The complexity of the tanh implementation of the original float tanh function is compared to that of the PWL approximation versions, detailed in Table~\ref{tab:fpga_resources}. For the implementation of the PWL approximation, no DSP slices are needed and the number of FF and LUT required is significantly reduced. Compared to the actual float tanh function, the 3-segment PWL requires 19 times less LUT, and 89 times less FF. As the number of segments increases, more resources are required.

In conclusion, when taking into account the performance, memory, and resources, the 3-segment PWL with re-training appears to be a viable candidate. The 3-segment PWL with re-training shows only a 0.1~dB drop in Q-factor from the model without the approximations. Memory usage is efficient as only a few coefficients must be stored. Additionally, the implementation of PWL does not require DSP slices, which is the most expensive block, resulting in more efficient use of available resources. Lastly, it is demonstrated that with the training used to reduce the approximation errors, we no longer observe a trade-off between the implementation complexity and the equalizer's performance.
\vspace{-2mm}



\begin{table}[h]
\centering
\vspace{-2mm}
\begin{tabular}{cccccc}
\hline
FPGA Resources & Original tanh & 3-Segment & 5-Segment & 7-Segment & 9-Segment \\ \hline\hline
\rowcolor[HTML]{EFEFEF} 
DSP Slices     & 26       & 0             & 0             & 0             & 0             \\
LUT            & 3849     & 203           & 570           & 1076          & 1374          \\
\rowcolor[HTML]{EFEFEF} 
FF             & 3020     & 34            & 98            & 164           & 230           \\ \hline
\end{tabular}
\caption{Tanh implementation complexity for PWL approximations for the FPGA realization.}
\label{tab:fpga_resources}
\end{table}
\vspace{-2mm}
\footnotesize
\textit{Acknowledgements}:
This work has received funding from the EU Horizon 2020 program under the Marie Sk\l{}odowska-Curie grant agreement No.~956713 (MENTOR) and 813144 (REAL-NET). SKT acknowledges the support of the EPSRC project TRANSNET (EP/R035342/1).
\vspace{-7mm}
\normalsize

\end{document}